\documentclass[sigconf]{acmart}

\AtBeginDocument{%
  }

\setcopyright{acmlicensed}
\copyrightyear{2026}
\acmYear{2026}
\acmDOI{XXXXXXX.XXXXXXX}
\acmConference[MM '26]{the 34th ACM International Conference on Multimedia}{Nov. 10--14, 2026}{Rio de Janeiro, Brazil}
\acmISBN{978-1-4503-XXXX-X/2026/11}

\usepackage{multirow}
\usepackage{graphicx}
\usepackage{booktabs}
\usepackage{enumitem}

\begin{document}

\title{Y-BotFrame: An Extensible Embodied Agent Framework for Quadruped Robot Assistants}

\author{Luyao Zhang}
\affiliation{%
  \institution{Xidian University}
  \city{Xi'an}
  \country{China}}
\email{lyzhang_5@stu.xidian.edu.cn}

\author{Ke Li}
\affiliation{%
  \institution{Xidian University}
  \city{Xi'an}
  \country{China}}
\email{like0413@stu.xidian.edu.cn}

\author{Yuan Ding}
\affiliation{%
  \institution{Xidian University}
  \city{Xi'an}
  \country{China}}
\email{25031110055@stu.xidian.edu.cn}

\author{Xulong Zhao}
\affiliation{%
  \institution{Xidian University}
  \city{Xi'an}
  \country{China}}
\email{xlzhao.apollo@stu.xidian.edu.cn}

\author{Guo Yu}
\affiliation{%
  \institution{Xidian University}
  \city{Xi'an}
  \country{China}}
\email{guoyu1999@stu.xidian.edu.cn}

\author{Chengwei Yan}
\affiliation{%
  \institution{Xidian University}
  \city{Xi'an}
  \country{China}}
\email{ycw@stu.xidian.edu.cn}

\author{Fuyu Dong}
\affiliation{%
  \institution{Xidian University}
  \city{Xi'an}
  \country{China}}
\email{fydong@stu.xidian.edu.cn}

\author{Jiawei Hu}
\affiliation{%
  \institution{Xidian University}
  \city{Xi'an}
  \country{China}}
\email{21009200286@stu.xidian.edu.cn}

\author{Di Wang}
\authornote{Corresponding author.}
\affiliation{%
  \institution{Xidian University}
  \city{Xi'an}
  \country{China}}
\email{wangdi@xidian.edu.cn}

\author{Nan Luo}
\affiliation{%
  \institution{Xidian University}
  \city{Xi'an}
  \country{China}}
\email{nluo@xidian.edu.cn}

\author{Gang Liu}
\affiliation{%
  \institution{Xidian University}
  \city{Xi'an}
  \country{China}}
\email{gliu@xidian.edu.cn}

\author{Quan Wang}
\affiliation{%
  \institution{Xidian University}
  \city{Xi'an}
  \country{China}}
\email{qwang@xidian.edu.cn}

\renewcommand{\shortauthors}{Zhang et al.}

\begin{abstract}
Quadruped robots are capable of traversing a wide range of complex terrains with high flexibility. As highly mobile ground-based intelligent platforms, they can be equipped with modules for navigation control, environmental perception, and intelligent interaction, thereby serving as real-world mobile deployment platforms for various algorithms. In this paper, we introduce Y-BotFrame, an extensible embodied platform that turns a robot into an intelligent ground assistant. Y-BotFrame integrates multimodal perception capabilities, including speech, vision, and LiDAR, and employs a large language model as the cognitive core for environmental understanding, contextual reasoning, and task planning. The system maps user natural-language instructions into executable embodied task units that can be carried out by the robot. Y-BotFrame supports natural interaction through voice commands and visual feedback, removing the need for a remote controller and enabling efficient human-robot collaboration. With a highly extensible framework, Y-BotFrame supports plug-and-play integration of new functional modules as well as modular upgrades and iterative development, offering a reference implementation for the real-world deployment of general-purpose, instruction-driven embodied agents.The supplementary video is available at \textcolor{blue}{https://xdei-group.github.io/Y-BotFrame/}
\end{abstract}

\ccsdesc[500]{Computing methodologies~Artificial intelligence}
\ccsdesc[500]{Computing methodologies~Planning and scheduling}

\keywords{Quadruped robots, embodied agents, large language models, multimodal interaction, autonomous navigation}

\maketitle

\section{Introduction}
\label{sec：intro}
Quadruped robots have demonstrated significant potential as ground-based mobile platforms with strong adaptability to complex terrains. By leveraging various gait patterns, they can achieve stable locomotion in unstructured environments such as stairs, gravel paths, and grasslands. Compared with conventional ground robots, quadruped robots exhibit stronger mobility, flexibility, and traversability in complex terrains, making them promising platforms for deploying general-purpose embodied agents. However, most existing intelligent systems for quadruped robots focus primarily on high-level path planning~\cite{hoeller2024anymal}, locomotion control~\cite{yu2021visual}, or task-specific policy learning~\cite{huang2025moe}. Some recent systems adopt vision-language-action models to formulate perception, planning, and control in an end-to-end manner~\cite{ding2024quar}. Although such methods can achieve promising performance in constrained scenarios, they usually depend on specific task distributions and large-scale training data, making it difficult to meet the requirements for multitask execution, extensibility, and stable deployment in open environments.

\begin{figure}[!t]
  \centering
  \includegraphics[width=\columnwidth]{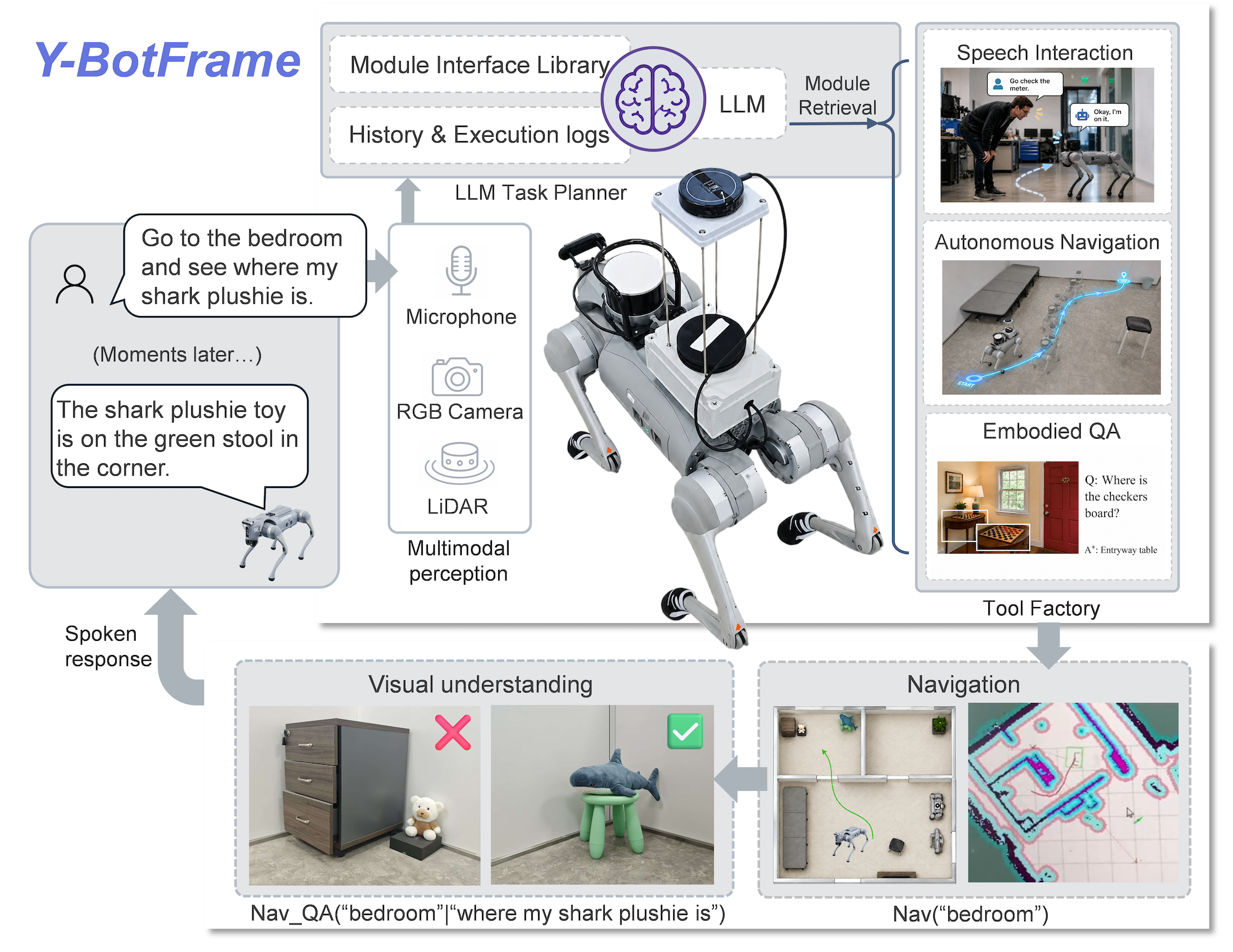}
  \caption{Overview of Y-BotFrame. The proposed system integrates multimodal human--robot interaction, an LLM-based task planner, executable tool modules, and robot-side execution feedback to support instruction-driven embodied tasks on a quadruped robot.}
  \Description{A system overview diagram of Y-BotFrame. The figure shows four main components: human--robot interaction and multimodal input, an LLM task planner, a tool factory with executable modules including speech interaction, autonomous navigation, and embodied question answering, and an execution and response module that returns task progress, exceptions, final results, and spoken responses.}
  \label{fig:Framework}
\end{figure}

In this paper, we propose Y-BotFrame, an embodied intelligent system that seamlessly integrates quadruped mobility, visual perception, and the semantic understanding and task planning capabilities of large language models. Y-BotFrame is designed as a hierarchical embodied platform for real-world deployment. Specifically, at the lower level, Y-BotFrame encapsulates capabilities such as navigation, environmental perception, and embodied question answering into executable modules with clearly defined input and output interfaces. At the upper level, an LLM-based agent~\cite{xi2025rise} serves as a task planner that analyzes user natural-language instructions, historical interaction records, and environmental priors to generate structured task plans. Through hierarchical decoupling and modular coordination, Y-BotFrame avoids the strong dependence of end-to-end models on large-scale task-specific data while improving the interpretability, extensibility, and stability of the system during real-world deployment.

\section{System Framework}
\label{sec:Sys}

\subsection{LLM Task Planner}
The task planner is implemented with a single large language model that maps user instructions to executable module interfaces. The system maintains structured descriptions of callable modules, including their functions, input parameters, applicable conditions, and execution constraints. Given a natural-language instruction, the LLM selects suitable modules based on user intent and binds the corresponding interfaces and parameters to generate a structured task plan. To ensure contextual consistency, the planner also incorporates historical plans and execution records, such as previous module invocations and feedback. Low-level coordination, real-time control, and safety constraints are handled by the execution modules, which continuously report task progress, exceptions, and final results to support monitoring and replanning.

\subsection{Tool Factory}
Y-BotFrame organizes its executable capabilities into a Tool Factory, which mainly includes speech interaction, autonomous navigation, and embodied question answering. These modules provide standardized interfaces for the LLM agent, allowing the system to translate user instructions into executable robot actions.

\noindent \textbf{Speech interaction.} The speech interaction module enables natural communication through ASR and TTS, while an LLM is used for dialogue understanding and response generation. To support domain-specific scenarios and long-term interaction, the module can further incorporate a knowledge base, voiceprint recognition, and dialogue memory.

\noindent \textbf{Autonomous navigation.} The navigation module provides localization, path planning, and safe motion for the quadruped robot. It fuses LiDAR, GNSS, and IMU measurements for global pose estimation, builds a local obstacle-aware grid map from LiDAR point clouds, and generates feasible paths using the A$^\ast$ algorithm. The map and planned path are updated online according to real-time perception, enabling navigation in complex environments without pre-built maps.

\noindent \textbf{Embodied question answering.} Based on these core capabilities, Y-BotFrame supports embodied question answering. After reaching the target location, the robot performs a rotational scan, captures multi-view RGB observations, and uses Qwen3-VL-Flash~\cite{bai2025qwen3} to generate scene descriptions. These descriptions, together with the user question and viewing angles, are then fused with an LLM to generate the final answer.

\subsection{System Workflow and Infrastructure}
The system supports natural human-robot interaction primarily through voice dialogue. Upon receiving a user request, the task planner first retrieves callable functions according to the currently available module interfaces and generates the corresponding module invocation instructions. If no suitable execution module is matched, the system parses the request as an open-ended dialogue task and returns a natural-language response to the user through speech synthesis.

Within this framework, the task planner invokes functional modules synchronously and receives asynchronous feedback over the same local area network. Modules can be deployed either on the robot or on a remote server according to computational load and real-time requirements. Heavy tasks such as model inference and visual understanding are handled on the server, while the robot retains lightweight perception, motion control, and result transmission. This distributed design reduces onboard computation and improves system extensibility, deployment flexibility, and operational efficiency.

\bibliographystyle{ACM-Reference-Format}
\bibliography{sample-base}

@String{Computer = "{IEEE} Computer" }

@String{Springer = "Springer-Verlag" }

@article{bai2025qwen3,
  title={Qwen3-vl technical report},
  author={Bai, Shuai and Cai, Yuxuan and Chen, Ruizhe and Chen, Keqin and Chen, Xionghui and Cheng, Zesen and Deng, Lianghao and Ding, Wei and Gao, Chang and Ge, Chunjiang and others},
  journal={arXiv preprint arXiv:2511.21631},
  year={2025}
}

@article{hoeller2024anymal,
  title={Anymal parkour: Learning agile navigation for quadrupedal robots},
  author={Hoeller, David and Rudin, Nikita and Sako, Dhionis and Hutter, Marco},
  journal={Science Robotics},
  volume={9},
  number={88},
  pages={eadi7566},
  year={2024},
  publisher={American Association for the Advancement of Science}
}

@inproceedings{yu2021visual,
  title={Visual-locomotion: Learning to walk on complex terrains with vision},
  author={Yu, Wenhao and Jain, Deepali and Escontrela, Alejandro and Iscen, Atil and Xu, Peng and Coumans, Erwin and Ha, Sehoon and Tan, Jie and Zhang, Tingnan},
  booktitle={5th Annual Conference on Robot Learning},
  year={2021}
}

@inproceedings{huang2025moe,
  title={Moe-loco: Mixture of experts for multitask locomotion},
  author={Huang, Runhan and Zhu, Shaoting and Du, Yilun and Zhao, Hang},
  booktitle={2025 IEEE/RSJ International Conference on Intelligent Robots and Systems (IROS)},
  pages={14218--14225},
  year={2025},
  organization={IEEE}
}

@inproceedings{ding2024quar,
  title={Quar-vla: Vision-language-action model for quadruped robots},
  author={Ding, Pengxiang and Zhao, Han and Zhang, Wenjie and Song, Wenxuan and Zhang, Min and Huang, Siteng and Yang, Ningxi and Wang, Donglin},
  booktitle={European Conference on Computer Vision},
  pages={352--367},
  year={2024},
  organization={Springer}
}

@article{xi2025rise,
  title={The rise and potential of large language model based agents: A survey},
  author={Xi, Zhiheng and Chen, Wenxiang and Guo, Xin and He, Wei and Ding, Yiwen and Hong, Boyang and Zhang, Ming and Wang, Junzhe and Jin, Senjie and Zhou, Enyu and others},
  journal={Science China Information Sciences},
  volume={68},
  number={2},
  pages={121101},
  year={2025},
  publisher={Springer}
}

\end{document}